\documentclass [10pt,twocolumn,letterpaper]{article}

\usepackage{iccv}
\usepackage{times}
\usepackage{epsfig}
\usepackage{graphicx}
\usepackage{amsmath}
\usepackage{amssymb}
\usepackage{mathrsfs}
\usepackage{tabularx}
\usepackage{authblk}
\iccvfinalcopy 

\def\iccvPaperID{2865} 

\ificcvfinal\fi

\begin{document}

\title{Occlusion Robust Face Recognition Based on Mask Learning \\ with Pairwise Differential Siamese Network}
\author[12]{Lingxue Song}
\author[1]{Dihong Gong}
\author[1]{Zhifeng Li\thanks{Corresponding author}}
\author[2]{Changsong Liu}
\author[1]{Wei Liu}
\affil[1]{Tencent AI Lab \qquad \qquad \qquad \qquad $^{2}$Tsinghua University}
\affil[ ]{\tt\small {songlx15@mails.tsinghua.edu.cn gongdihong@gmail.com michaelzfli@tencent.com lcs@tsinghua.edu.cn wl2223@columbia.edu}}

\renewcommand\Authands{ and }

\maketitle
\ificcvfinal\thispagestyle{empty}\fi

\begin{abstract}
Deep Convolutional Neural Networks (CNNs) have been pushing the frontier of the face recognition research in the past years. However, existing general CNN face models generalize poorly to the scenario of occlusions on variable facial areas. Inspired by the fact that a human visual system explicitly ignores occlusions and only focuses on non-occluded facial areas, we propose a mask learning strategy to find and discard the corrupted feature elements for face recognition. A mask dictionary is firstly established by exploiting the differences between the top convoluted features of occluded and occlusion-free face pairs using an innovatively designed Pairwise Differential Siamese Network (PDSN). Each item of this dictionary captures the correspondence between occluded facial areas and corrupted feature elements, which is named Feature Discarding Mask (FDM). When dealing with a face image with random partial occlusions, we generate its FDM by combining relevant dictionary items and then multiply it with the original features to eliminate those corrupted feature elements. Comprehensive experiments on both synthesized and realistic occluded face datasets show that the proposed approach significantly outperforms the state-of-the-arts.
\end{abstract}
\section{Introduction}
Deep Convolutional Neural Networks (CNNs) have recently made a remarkable improvement in unconstrained face recognition problem. Researchers are racing in ways to boost the performance using advanced network architectures \cite{VGG,Resnet,Googlenet,SEnet,wang2018orthogonal} or designing new loss functions to facilitate discriminative feature learning \cite{Tripletloss,centerloss,sphereface,cosface,arcface,zhang2017range, wang2019decorrelated,yang2019face,dong2019efficient}. Some of them even surpass human recognition ability on certain benchmark database \cite{lfw}.

Despite the huge success of deep learning models under general face recognition scenario, the deep features still show imperfect invariance to uncontrollable variations like pose, facial expression, illumination, and occlusion. Among all these factors, occlusion has been considered a highly challenging one. In real-life images or videos, facial occlusions can often be observed, \eg facial accessories including sunglasses, scarves, and masks or other random objects like books and cups. As indicated in~\cite{CVPR2015Analysis}, without specifically trained with a large number of occluded face images, deep CNN-based models indeed cannot function well because of the larger intra-class variation and higher inter-class similarity that caused by occlusions.

\vspace{-0.1em}
One possible approach to improve the performance of CNN models under partial occlusions is to train the network with occluded faces. Daniel \etal~\cite{Enhancing} proposed to augment training data with synthetic occluded faces in a strategic manner and observed improved performance. However, it does not solve the problem intrinsically because it only ensures the features are more locally and equally extracted, as analyzed in~\cite{increasing}. The inconsistency between features of two faces with different occlusion situations still exists. For example, features of an occlusion-free face bear much more information in eyes area than that of a face wearing a pair of sunglasses unless the network is trained not to utilize the eyes area at all, which is unreasonable.

\begin{figure*}[t]
\begin{center}
\includegraphics[width=0.85\linewidth, height=3.0in]{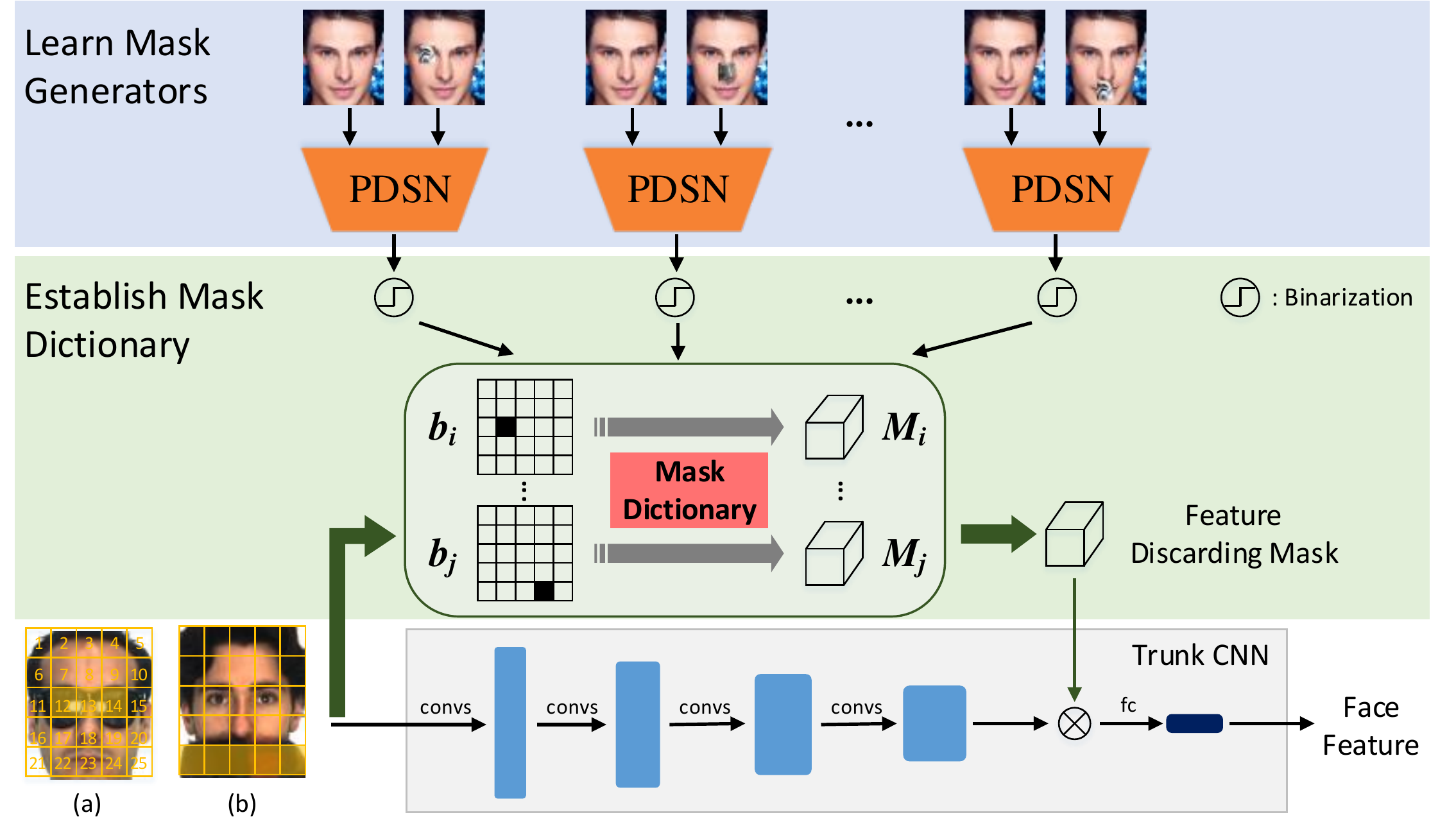}
\end{center}
\caption{An overview of the proposed framework. Based on a trunk CNN model trained for face recognition, we propose the pairwise differential siamese network (PDSN) structure to learn the correspondence between occluded facial blocks and corrupted feature elements. Then a mask dictionary is established accordingly, which is used to composite the feature discarding mask (FDM) for a test face with random partial occlusions. Finally, we multiply the FDM with original face features to eliminate corrupted feature elements from recognition.}
\label{fig:figure1}
\end{figure*}
Inspired by the fact that the human visual system pays attention to the non-occluded facial areas for recognition (and ignores the occluded areas), we propose to discard feature elements that have been corrupted by occlusions. A core question would be: \emph{given a face image with random partial occlusions, how to locate those corrupted feature elements?} It is not a big deal for traditional low-level features like LBP, HOG or SIFT because there is a clear correspondence between image pixels and final feature elements, but what about the deep CNN features? Therefore, the core of this work is to find corrupted feature elements under random partial occlusion and eliminate the response of these elements from the recognition process. It is worth stated that the facial occlusion detection problem is not the focus of this paper, thus we directly adopt a similar way as~\cite{FacialSeg} to detect the occlusion location in image space. 

To learn the correspondence between occluded facial regions and corrupted feature elements, we develop a novel pairwise differential siamese network (PDSN) structure with a mask generator module that takes pairwise images (one is a clean face and the other is an occluded one of the same identity) as input. The differential signal between the conv features of clean and corresponding occluded faces is fed into the mask generator module. It acts as a role of attention mechanism which encourages the model to focus on those feature elements that have deviated from its true values owing to partial occlusions. Moreover, we propose to learn the mask generator by minimizing a combination of two losses: a pairwise contrastive loss that penalizes the large differences between the masked conv features of clean and occluded faces, and a classification loss which ensures those feature elements that harm the recognition are masked out. With these two losses, our mask generator will identify those feature elements that are harmful for the recognition as well as far from its genuine values as corrupted ones. To handle the random partial occlusions, we first divide the aligned face into several predefined blocks and only learn PDSNs for these blocks, since severely performance dropping usually only occurs when critical facial components are missing. Then we construct a mask dictionary from these trained PDSNs by strategic binarization. Each item in this dictionary is a binary mask, named Feature Discarding Mask (FDM), which indicates the feature elements that should be set to zero when one facial block is occluded. In the testing phase, the FDM of a face with random partial occlusions is derived by element-wise logical ``AND" of relevant dictionary items, which is then multiplied with the original face feature to discard those corrupted feature elements from the recognition. Figure~\ref{fig:figure1} gives an overview of the proposed framework.

The main contributions of this paper are two-fold: (1) we propose a novel PDSN framework to explicitly find correspondence between occluded facial blocks and corrupted feature elements for deep CNN models, which is innovative and inspiring; (2) based on the PDSN, we develop a face recognition system that is robust for occlusions. Our system demonstrates superior performance on face datasets with both realistic and synthesized occlusions and generalizes very well on general face recognition tasks.
\section{Prior Work}

Partial occlusion is one of the major challenges in face recognition that has received much attention in the era of hand-crafted features. Before the emergence of deep CNNs, face recognition under partial occlusions has been typically handled using two types of algorithms, namely, (i) methods that extract local face descriptors only from the non-occluded facial areas or (ii) methods that recover clean faces from the occluded ones. The first type usually explicitly divides face images into several local regions. A support vector machine (SVM) is trained to identify which local regions are occluded and then they are discarded from recognition~\cite{localnon-neg,improving,Park2015Partially}, with optional subspace methods \cite{li2005nonparametric, liu2006spatio} to reduce feature dimension before the classification stage. However, the discriminative power of this kind of approach is limited in view of using shallow features like Local Gabor Binary Patterns (LGBP)~\cite{localnon-neg}. Among the second type of methods, the sparse representation-based classification (SRC)~\cite{SRC} is considered to be the pioneering work on occlusion robust face recognition. This model reconstructs an occlusion-free face using a linear combination of images from the training set together with a sparse constraint term accounting for occlusions. Inspired by this model, researchers extended it by rethinking the distribution of the sparse constraint term~\cite{RSC,correntropy,CESR} or characterizing the structure information of it~\cite{MRF,SSEC}. These approaches do not generalize well since they require test samples have identical subjects with the training samples. 
\begin{figure}[t]
\centering
\begin{minipage}{0.45\linewidth}
    \includegraphics[width=3.25cm]{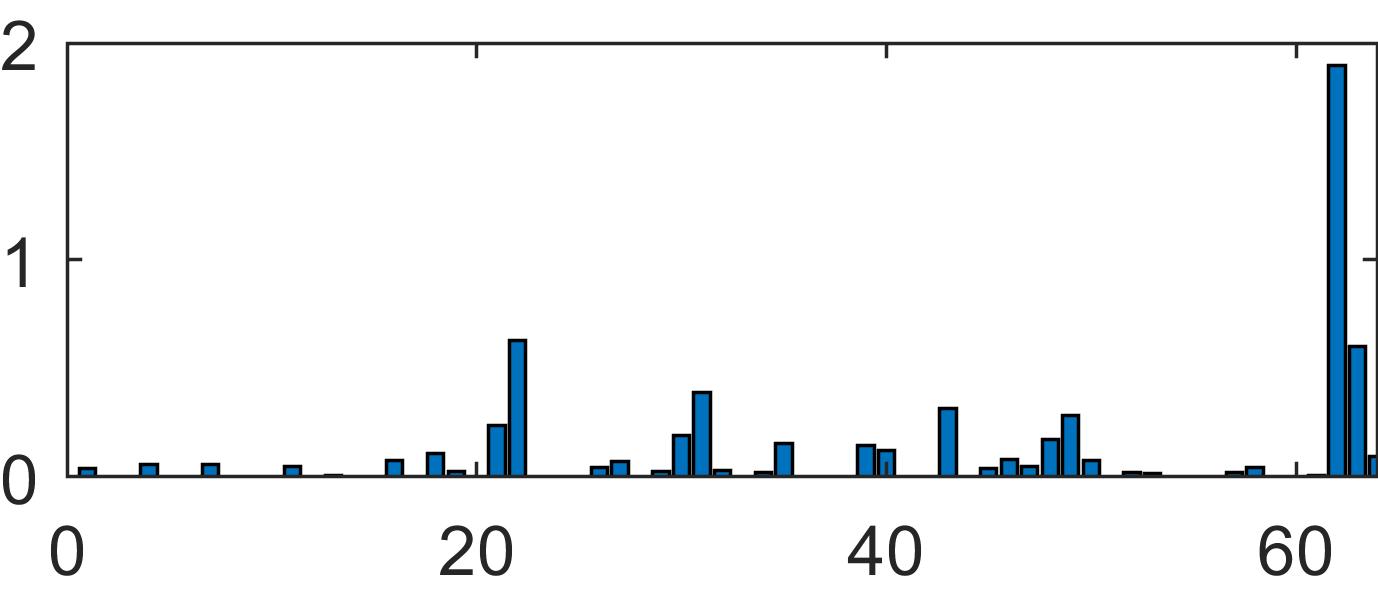}
\end{minipage}
\begin{minipage}{0.45\linewidth}
    \includegraphics[width=3.25cm]{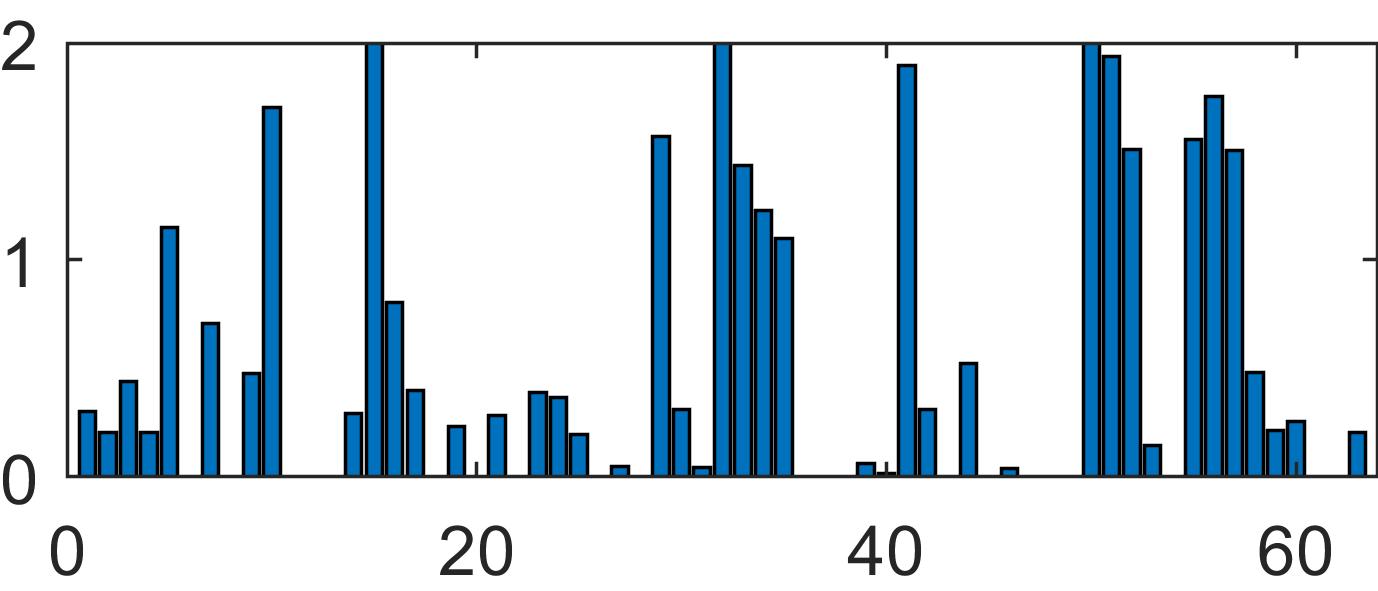}
\end{minipage} \\[2pt]

\begin{minipage}{0.45\linewidth}
    \includegraphics[width=3.25cm]{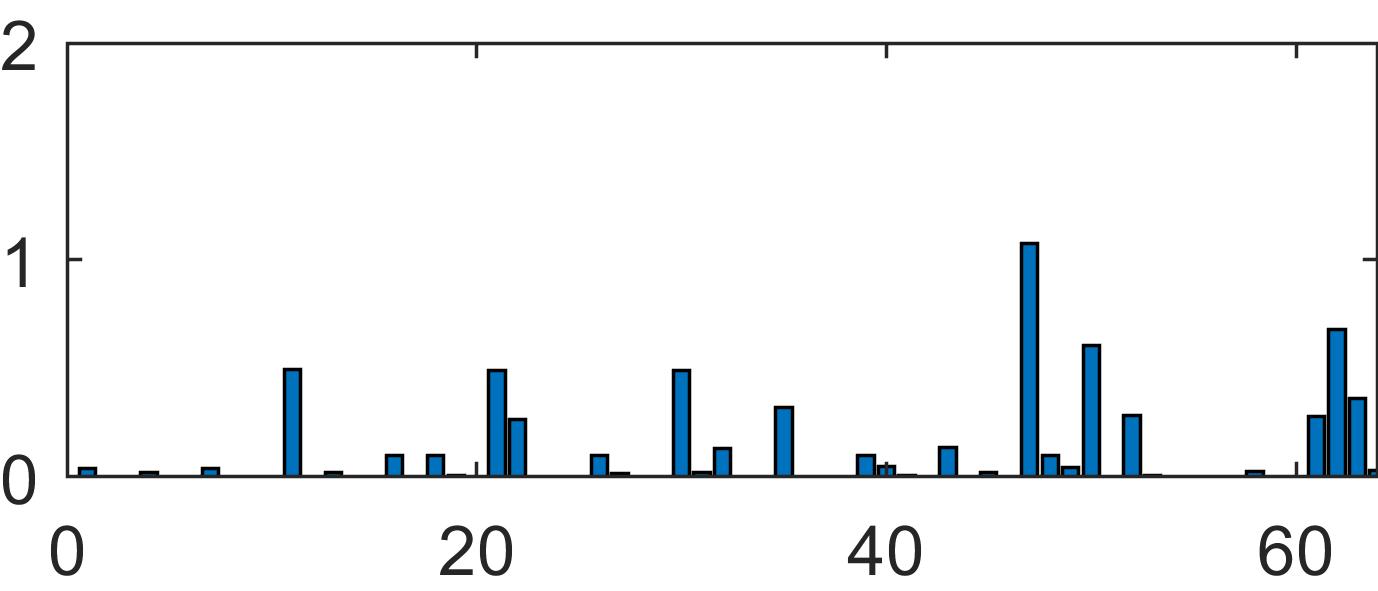} 
\end{minipage}
\begin{minipage}{0.45\linewidth}
    \includegraphics[width=3.25cm]{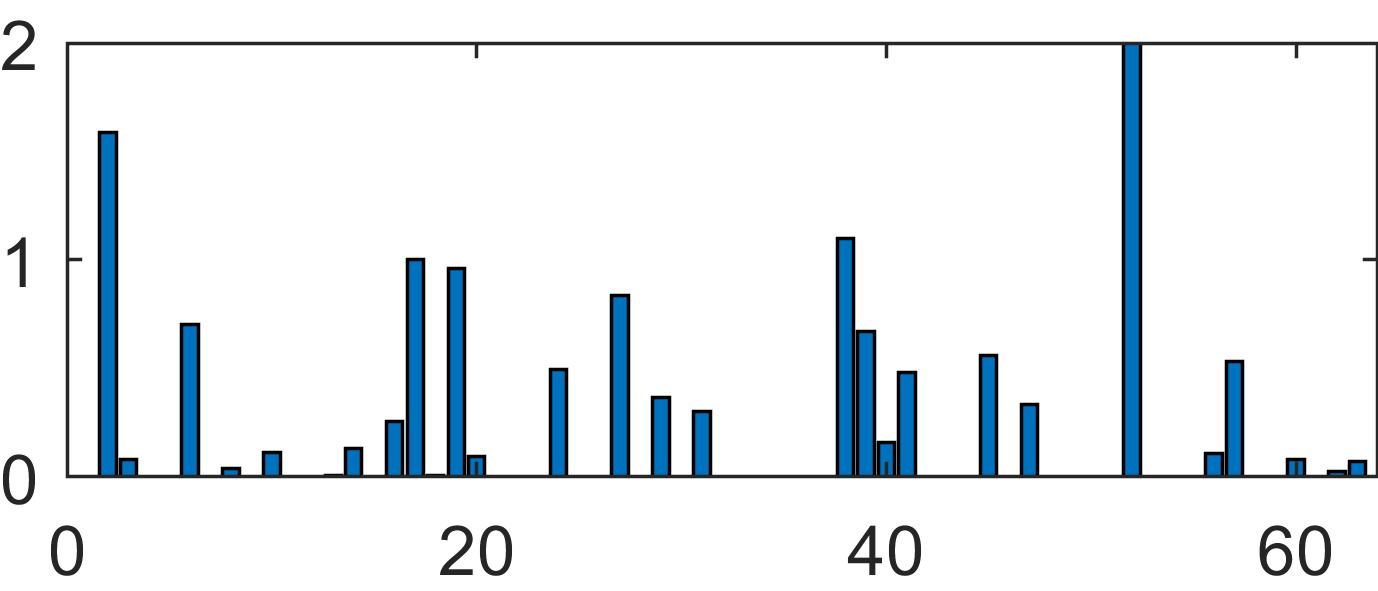}
\end{minipage}\\[10pt]
\caption{Neural response differences between two face images of different subjects with the same partial occlusion. Left: neural activation differences of the top conv layer. Right: neural activation differences of the top fc layer. We randomly sample 64 neurons for illustration here.}
\label{fig:figure2}
\end{figure}

Deep learning has been dominant in the field of face recognition for several years. As early as 2014, Sun \etal~\cite{DeepID2+} have discovered that the feature learned by DeepID2+ show certain degree of robustness to image corruption in face verification task. Combining DeepID2+ features extracted from 25 face patches further improves the robustness. Cheng \etal~\cite{LSTMAuto} present an LSTM-Autoencoder to restore occluded facial areas in the wild and carried out recognition on the recovered face images. But there is no guarantee the recovered part indeed matches the identity of the individuals to be recognized especially under the open-set scenario. Daniel \etal~\cite{Enhancing} tackle the occlusion problem by augmenting training data with synthetic occluded faces that only specific regions where a CNN extracts the most discriminative features from are covered. In this way, features are more equally and locally extracted. Wan \etal~\cite{MaskLearning} propose to add a MaskNet branch to the middle layer of CNN models which is expected to assign lower weights to hidden units activated by the occluded facial areas. But the middle conv layer is not discriminative enough and the MaskNet branch lacks additional supervision information to ensure the functionality. 

In a word, the discriminative ability of traditional low-level feature-based methods is limited, and the existing few deep learning-based methods lack the awareness of how partial occlusions truly affect the CNN models. The inconsistency between features of two faces with different occlusion situations has not been carefully considered yet. The proposed method complements the missing piece of the puzzle and is able to explicitly locate corrupted feature elements for trained CNN models and discard them from the recognition, to ensure a fair comparison. Therefore, our approach is an intrinsic way with good generalization ability compared to the aforementioned studies.

\section{Proposed Approach}

The overall pipeline of the proposed approach is shown in Figure~\ref{fig:figure1}, which decomposes the problem of face recognition under random partial occlusions into three stages. \emph{Stage} \uppercase\expandafter{\romannumeral1}: Learning mask generators using the proposed pairwise differential siamese network (PDSN) to capture the correspondence between occluded facial blocks and corrupted feature elements. \emph{Stage} \uppercase\expandafter{\romannumeral2}:  Establishing a mask dictionary from the learned mask generators. \emph{Stage} \uppercase\expandafter{\romannumeral3}:  In the testing phase, combining the feature discarding mask (FDM) of random partial occlusions from this dictionary, which is then multiplied with the original feature to eliminate the effect of partial occlusions from recognition. 

\subsection{Stage \uppercase\expandafter{\romannumeral1}: Learning Mask Generators}
\subsubsection{Problem Analysis}
\label{sec:discuss}

\quad Face images fed into the CNN model are mostly well-aligned by detected facial keypoints, we divide the aligned face into non-overlapping $N \times N$ blocks, denoted as $\{b_{j}\}^{N*N}_{j=1}$, and aim to learn a mask generator for every $b_{j}$ to find the corrupted feature elements when this block is occluded. In our implementation, we set $N=5$ according to the input image size so that the facial components like eyes, nose tip and mouth are appropriately associated with a block. The face (a) in Figure~\ref{fig:figure1} gives the division example.

Then we define our core problem in Stage \uppercase\expandafter{\romannumeral1} as: given the feature of a face image with block $b_{j}$ occluded, denoted as $f(x_{j})$, how to learn a mask generator $\mathscr{M}_{\theta}$ whose output is multiplied with the $f(x_{j})$ to mask out those corrupted elements. Let the purified feature be denoted as $\Tilde{f}(x_{j})$, then $\Tilde{f}(x_{j})=\mathscr{M}_{\theta}(\cdot)f(x_{j})$. There are two choices to be decided before running into the learning process:
\begin{figure}[t]
\begin{center}
\includegraphics[width=0.98\linewidth]{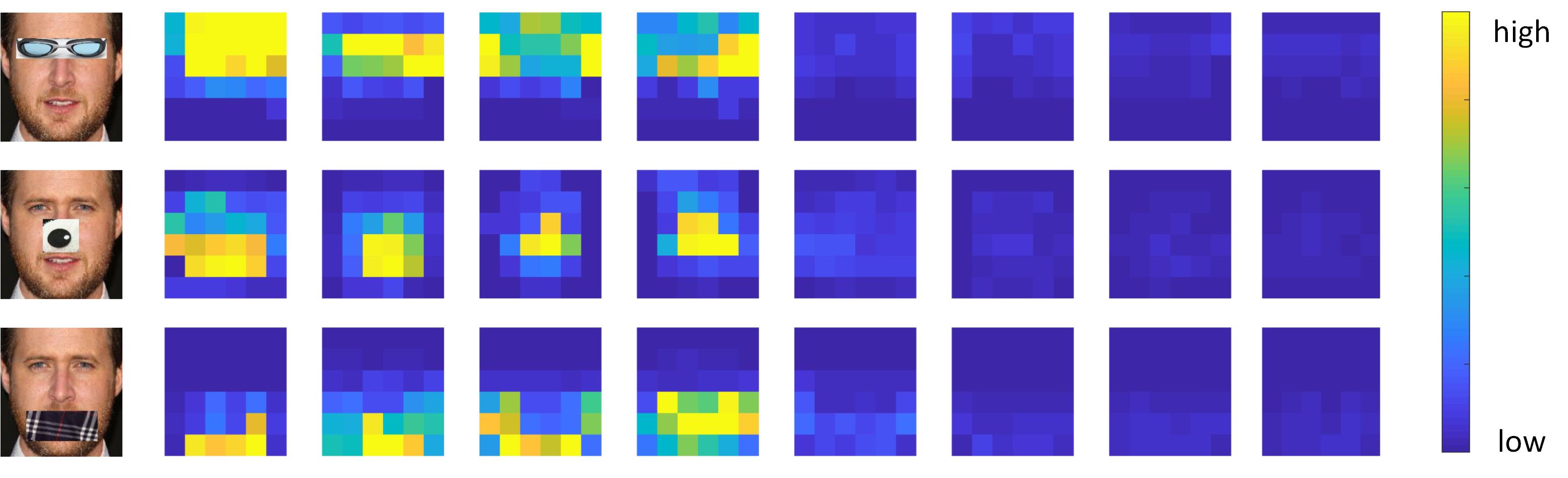} \\[0pt]
\end{center}
   \caption{The median relative rate of change (MED) of neuron's activation values in the top conv layer under three types of occlusions. We select eight channels for illustration here.}
\label{fig:figure3}
\end{figure}

\textbf{The choice of $\textbf{f}$}. For the CNN-based face recognition model, the face feature usually refers to the output of the final fully-connected (fc) layer before the classification layer. However, every neuron in the fc layer integrates
information from all the output elements of the previous layer, so the occluded areas might be mixed up with the non-occluded areas in the final fc feature. From another perspective, neurons in the top fc layers are highly selective to identities~\cite{DeepID2+}. Therefore even if different subjects are contaminated by the same occlusion, the positions of feature elements that changed by this occlusion will be highly dependant on face identity, as shown in the rightmost column in Figure~\ref{fig:figure2}. In contrast, we can see from the left column of Figure~\ref{fig:figure2} that the positions of feature elements that changed by the same occlusion of different individuals are quite consistent for the top conv layer, and it still preserves local information, thus we choose the top conv feature as our $f$.

\textbf{The output dimension of $\mathscr{M}_{\theta}$}. In~\cite{MaskLearning}, they learned a 2D mask $M \in \mathbb{R}^{W \times H}$ for the 3D conv feature maps $U \in \mathbb{R}^{C \times W \times H}$. That is to say, the feature elements of all $C$ channels in the same spatial location share the same weight from their learned mask. In other words, they assumed that feature elements of all the conv feature channels respond the same to the occlusion. With questions about the rationality of their hypothesis, we'd like to gain more insights into the true reaction of the top conv feature to partial occlusions. We use a criterion named \emph{median relative rate of change} (denoted as MED) to capture the extent to which each feature element is away from its true value under partial occlusions. Given a pair of clean face image $x_{clean}$ and its corresponding occluded face image $x_{occ}$, we first calculate the \emph{relative rate of change} of neuron activation values of the top conv layer:
\begin{equation}
r_{i} = |\frac{f^i(x_{clean}) - f^i(x_{occ})}{f^i(x_{clean})}| 
\label{equation1}
\end{equation}
where $r_{i}$ denotes the \emph{relative rate of change} of the $i^{th}$ feature element value of the top conv layer. 
We randomly select N images from the CASIA-WebFace~\cite{CASIA-WebFace} and add occlusions on the faces, then calculate the $r_{i}$s for every face pair. The metric MED to approximately represents the altered degree of the $i^{th}$ feature element under occlusions is obtained by calculating the median value of these $r_{i}$s. If the MED of a feature element is high when an area of the input face is occluded, then it will likely bring unreasonable noise into the final feature. 

In Figure~\ref{fig:figure3} we show the MED values of feature elements in 8 channels of the top conv feature maps under three types of occlusions. Obviously, the feature values are altered in a different way for different channels, elements of some channels change very little while elements of some channels change drastically in the same spatial locations. This is interesting because in view of the receptive field, the same spatial location of different conv channels gather information from the same region of the input image, but they actually react quite differently to occlusions. Therefore, we believe the output dimension of $\mathscr{M}_{\theta}$ should be the same as the top conv feature maps, which is $C \times W \times H$.
\begin{figure*}[t]
\begin{center}
\includegraphics[width=0.85\linewidth]{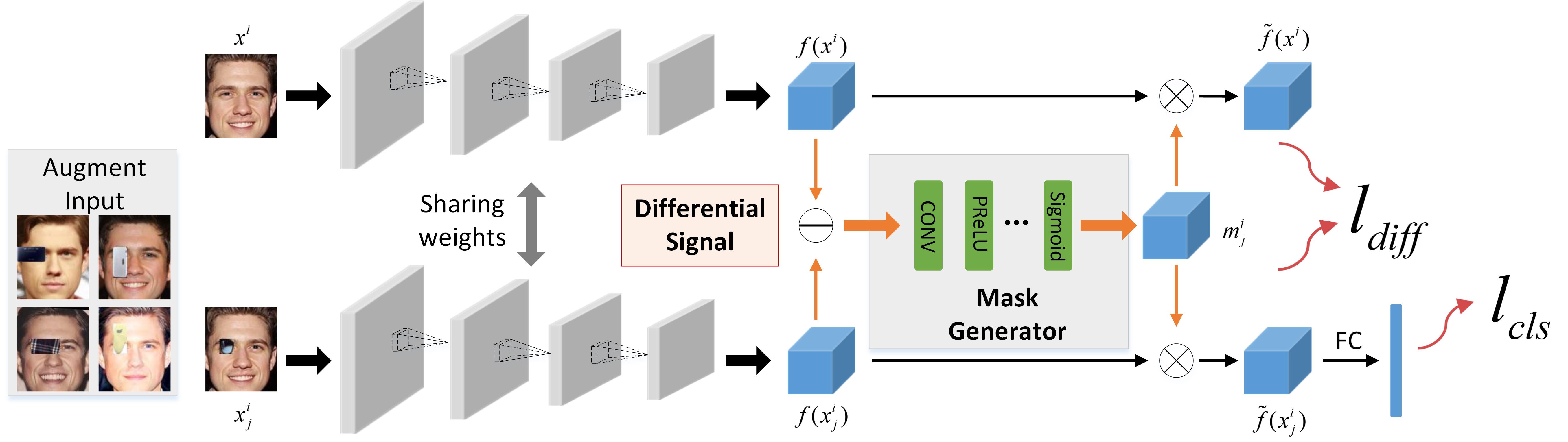}
\end{center}
\caption{The illustration of the proposed Pairwise Differential Siamese Network.}
\label{fig:figure4}
\end{figure*}
\vspace{-0.5em}
\subsubsection{Pairwise Differential Siamese Network}

\quad Given the analysis in Sec.~\ref{sec:discuss}, we propose the pairwise differential siamese network (PDSN) structure to learn the relations between occluded facial blocks and corrupted feature elements. As illustrated in Figure~\ref{fig:figure4}, it consists of a trunk CNN and a mask generator branch, forming a siamese architecture. The trunk CNN is responsible for extracting base face representation, which is shared by the clean and occluded face pairs and could be any CNN architecture. The core mask generator module $\mathscr{M}_{\theta}$ in our PDSN is expected to output a mask whose element is a real value in $[0,1]$ and is multiplied with the input contaminated feature to diminish its corrupted elements: $\Tilde{f}(x^i_{j}) = \mathscr{M}_{\theta}(\cdot)f(x^i_{j})$, where $f(\cdot)$ is top conv feature and $x^i_{j}$ denotes occluded face image of the $i^{th}$ pair. The two faces inside an input pair belong to the same identity $y^i$ and the only difference is that one of them has partial occlusion on the facial block $b_{j}$. The key requirement for learning the mask generator is that the remaining part of the feature $f(x^i_{j})$ after masking should be as similar to its corresponding clean feature  $f(x^i)$ as possible while guarantees a successful recognition.

To this end, we propose to learn $\mathscr{M}_{\theta}$ by minimizing a combination of two losses:
\begin{equation}
L_{\theta} = \sum_{i}\ell_{cls}(\theta;\Tilde{f}(x^i_{j}), y^i) + \lambda \ell_{diff}(\theta;\Tilde{f}(x^i_{j}),\Tilde{f}(x^i)) 
\label{equation2}
\end{equation}
The first part of the cost, $\ell_{cls}$, is accounting for evaluating the importance of each feature element for recognition, and the second part, $\ell_{diff}$, assesses how far the feature of an occluded face is away from its true value. We will expand this formulation in the following part.

\textbf{The classification loss $\ell_{cls}$}. To find the corrupted feature elements, an intuitive idea is that, these feature elements contribute little to identifying the input face and may instead cause higher classification loss. Therefore the most straightforward supervision signal is the identity information, that is, the occluded face should be correctly classified by the classifier of the trunk CNN after masking, which gives us the first loss item (softmax loss for example):
\begin{equation}
\ell_{cls}(\theta;\Tilde{f}(x^i_{j}), y^i) = -log(p_{y^i}(F(\Tilde{f}(x^i_{j}))))
\label{equation3}
\end{equation}
The $\Tilde{f}(x^i_{j})$ is the top conv feature of an occluded face after masking, $F$ is the fc layer(s) of the trunk CNN model next to the top conv layer, and it could also be the average pooling layer in models like~\cite{sphereface}.

\textbf{The differential signal and pairwise loss $\ell_{diff}$}. The results shown in Figure~\ref{fig:figure3} inspired us that the differential signal between the top conv activation values of an occluded face and its corresponding clean one could be a good indicator of which feature elements are potential corrupted ones. To put it another way, the differential input signal acts as a role of attention mechanism which encourages the mask generator to focus on those feature elements that have deviated from its true values owing to partial occlusion. Therefore we feed our mask generator module with the absolute difference between features of an occlusion-free face and its occluded counterpart.

To further make use of the supervision information of what this subject's occlusion-free feature looks like, we propose a pairwise contrastive loss that minimizes per-element differences between the masked features of the occluded and occlusion-free faces:
\begin{multline}
\ell_{diff}(\theta;\Tilde{f}(x^i_{j}),\Tilde{f}(x^i))  \\
= \lVert \mathscr{M}_{\theta}(\cdot)f(x^i) -  \mathscr{M}_{\theta}(\cdot)f(x^i_{j})\rVert_1
\label{equation4}
\end{multline}
where $\mathscr{M}_{\theta}(\cdot)=\mathscr{M}_{\theta}(|f(x^i_{j}) - f(x^i)|)$, and $\lVert \cdot \rVert_1$ is the L1 norm. Obviously, this contrastive loss will punish those feature elements of the occluded face which are largely different from its occlusion-free one. Together with the classification loss, our mask generator will identify those feature elements that are harmful for the recognition as well as far from its genuine values as corrupted ones.

Thus, the overall object function in Eq.~\eqref{equation2} used in our implementation is:
\begin{equation}
\begin{split}
L_{\theta} = -\sum_{i}log(p_{y_i}(F(\mathscr{M}_{\theta}(\cdot)f(x^i_{j}))))  \\ 
+ \lambda \lVert \mathscr{M}_{\theta}(\cdot)f(x^i) -  \mathscr{M}_{\theta}(\cdot)f(x^i_{j})\rVert_1
\label{equation5}
\end{split}
\end{equation}
The $\lambda$ is set to 10 to make the different components of the object function have the same scale in our experiments.

We implement $\mathscr{M}_{\theta}$ as a module with several conv blocks and learn the different $\theta$ for occlusions on different facial blocks. The different $\theta$ is accounting for the distinct property of different facial components. For example, the eyes bear much more significance than the cheek area, therefore the input distribution of the mask generator varies accordingly. When learning mask generator $j$, in addition to the faces that only the target block $b_j$ is occluded, we augment samples with other blocks also occluded, which are the 4-neighbors of the target block $b_j$, to capture the dependency of adjacent blocks, as shown in Figure~\ref{fig:figure4}.
\subsection{Stage \uppercase\expandafter{\romannumeral2}: Establishing the Mask Dictionary}
In the testing phase, we don't have the paired images of a probe face and its occlusion location is random. Therefore, the trained PDSNs cannot be directly used to output the feature discarding mask(FDM) of a probe face. In {Stage \uppercase\expandafter{\romannumeral2}}, we would like to extract a fixed mask from every trained mask generator $\mathscr{M}_{\theta}$ and build a dictionary accordingly.

For a mask generator $\mathscr{M}_{\theta_j}$, we first feed the trained network with large amount of face pairs, one of which is occluded on the $j^{th}$ facial block and obtain the output masks of this generator, forming a large set of ${m^1_j, m^2_j,\dots, m^P_j}$, where $P$ (about 200k in our experiment) is the number of the face pairs. After Min-Max normalizing each $m^i_j$, we calculate the element-wise mean value of these $m^i_j$s and get a mean mask $\bar{m}_j$. It's possible to directly use this $\bar{m}_j$ as the FDM when the $j^{th}$ block is occluded (referred to as \textit{soft weight} schema). But this will reserve feature elements with very low mask values, which is inappropriate since the facial components inside this block have been totally lost. Therefore we believe setting those feature elements to zero to completely remove the noise is critical. We'll validate this in Sec.~\ref{sec:ablation}. The binarized FDM $M_j \in \mathbb{R}^{C \times W \times H}$ for this mask generator is derived by setting the feature locations with the smallest top $\tau*K$ mean values to zero:
\begin{equation} M_{j}[k] =
\begin{cases}
0 & \text{if  }  \bar{m}_j[k] \in \{\tilde{m}_j[1], \dots ,\tilde{m}_j[\tau*K]\},\\
1 & \text{else}.
\end{cases}
\label{equation6}
\end{equation}
where $k=1,2, \ldots ,K$, $K = C \times W \times H$ , $k$ denotes the feature index, and $\{\tilde{m}_j[1], \dots ,\tilde{m}_j[\tau*K]\}$ is the sorted smallest $\tau*K$ values of $\bar{m}_j$. $\tau$ is the discarding threshold and it will be discussed later in Sec.~\ref{sec:ablation}. By this way, we construct a mask dictionary that every item is a binary mask which indicates whether to discard each feature element when one certain block of the aligned face is occluded. 
\subsection{Stage \uppercase\expandafter{\romannumeral3}: Occlusion Robust Recognition}
\label{sec:algorithm}
\begin{figure}[t]
\begin{center}
\includegraphics[width=0.98\linewidth, height = 0.82in]{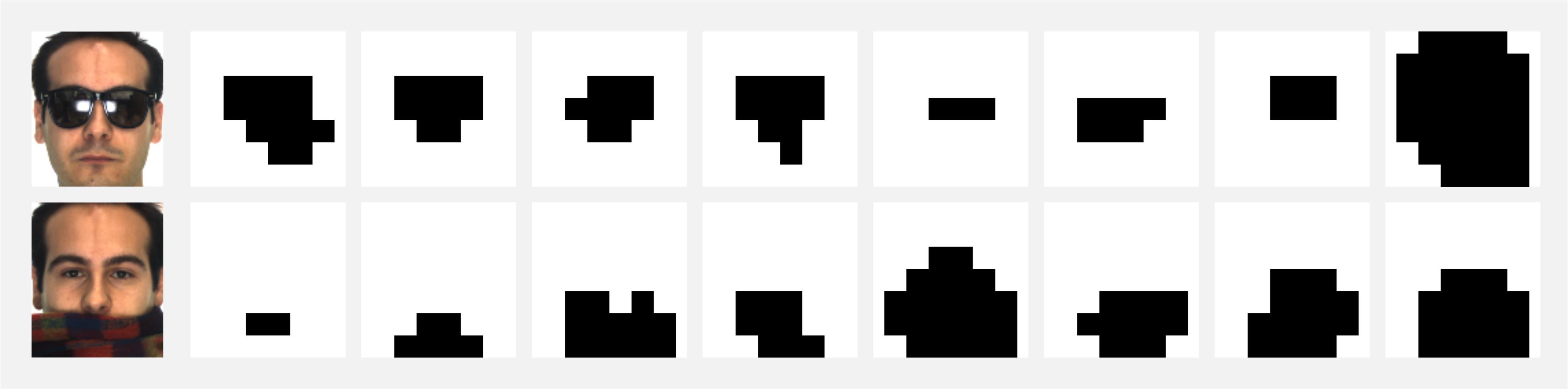} \\[0pt]
\end{center}
   \caption{Examples of the feature discarding masks of two occlusion types combined from our mask dictionary.}
\label{fig:figure5}
\end{figure}

With this mask dictionary, the FDM of a face with arbitrary partial occlusions could be derived by combining relevant dictionary items. By relevant we mean that if the occlusion area in a probe face has at least 0.5 IoU with a predefined facial block from the dictionary, we count this block as an occluded one for this face. For example, for the face (a) wearing sunglasses in Figure~\ref{fig:figure1}, its occlusion region covers block $\{b_{j}\}_{j=12}^{14}$, therefore its FDM is calculated by $M = M_{12} \land M_{13} \land  M_{14}$, where $\land$ denotes the element-wise logical ``AND" and the result $M$ is still a binary mask. Figure~\ref{fig:figure5} shows 8 channels of the FDM composited from the dictionary for sunglasses and scarf occlusions respectively.
\section{Experiments}
\label{sec:experiments}
\subsection{Implementation Details}
\label{sec:settings}

\noindent \textbf{Preprocessing.} The standard MTCNN~\cite{MTCNN} is used to detect 5 face landmarks for all the images. After performing similarity transformation accordingly, we obtain the aligned face images and resize them to be $112\times 96$ pixels. 

\begin{figure}[t]
\centering
\begin{minipage}{0.11\linewidth}
    \includegraphics[width=0.93cm]{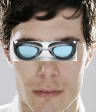}
\end{minipage}
\hfill
\begin{minipage}{0.11\linewidth}
    \includegraphics[width=0.93cm]{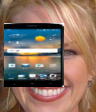}
\end{minipage}
\hfill
\begin{minipage}{0.11\linewidth}
    \includegraphics[width=0.93cm]{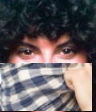}
\end{minipage}
\hfill
\begin{minipage}{0.11\linewidth}
    \includegraphics[width=0.93cm]{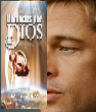}
\end{minipage}
\hfill
\begin{minipage}{0.11\linewidth}
    \includegraphics[width=0.93cm]{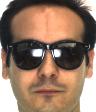}
\end{minipage}
\hfill
\begin{minipage}{0.11\linewidth}
    \includegraphics[width=0.93cm]{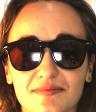}
\end{minipage}
\hfill
\begin{minipage}{0.11\linewidth}
    \includegraphics[width=0.93cm]{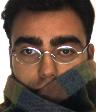}
\end{minipage}
\hfill
\begin{minipage}{0.11\linewidth}
    \includegraphics[width=0.93cm]{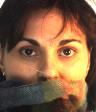}
\end{minipage}\\[2pt]

\begin{minipage}{0.11\linewidth}
    \includegraphics[width=0.93cm]{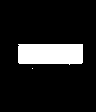}
\end{minipage}
\hfill
\begin{minipage}{0.11\linewidth}
    \includegraphics[width=0.93cm]{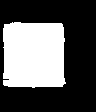}
\end{minipage}
\hfill
\begin{minipage}{0.11\linewidth}
    \includegraphics[width=0.93cm]{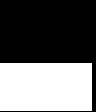}
\end{minipage}
\hfill
\begin{minipage}{0.11\linewidth}
    \includegraphics[width=0.93cm]{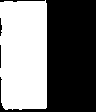}
\end{minipage}
\hfill
\begin{minipage}{0.11\linewidth}
    \includegraphics[width=0.93cm]{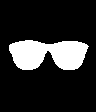}
\end{minipage}
\hfill
\begin{minipage}{0.11\linewidth}
    \includegraphics[width=0.93cm]{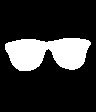} 
\end{minipage}
\hfill
\begin{minipage}{0.11\linewidth}
    \includegraphics[width=0.93cm]{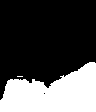}
\end{minipage}
\hfill
\begin{minipage}{0.11\linewidth}
    \includegraphics[width=0.93cm]{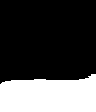}
\end{minipage}\\[8pt]
\caption{Occlusion detection results of our FCN-8s segmentation network on the occluded Facesrub and AR test images.}
\label{fig:figure6}
\end{figure}

\noindent \textbf{Occlusion Detection.} We train an FCN-8s~\cite{fcn} segmentation network to detect the occlusion location. The training data includes the synthetic occluded CASIA-WebFace dataset and images of 26 subjects (outside the test subjects) from the AR dataset. The vgg16 backbone is firstly trained with sufficient face images to provide a good initialization. Finally, our occlusion detection model works pretty well with a mean IU of 98.51 on our synthetic occluded Facescrub dataset~\cite{facescrub}. Figure~\ref{fig:figure6} shows some detection results. 

\noindent \textbf{Network Structure.} We employ the refined ResNet50 model proposed in recently published ArcFace~\cite{arcface} as our trunk CNN model. The mask generator is simply implemented as a CONV-PReLU-BN structure with a sigmoid function to map the output into $[0,1]$.

\noindent \textbf{Training.} The training procedure includes three stages. \emph{Stage} 1: Train the trunk CNN on the CASIA-WebFace~\cite{CASIA-WebFace} dataset with the large margin cosine loss~\cite{cosface}. \emph{Stage} 2: Fixing the model parameters of the trunk CNN, and train the mask generator modules with specifically designed face pairs as shown in Figure~\ref{fig:figure4}. We discovered that occlusions on the peripheral blocks of the faces barely affect the recognition accuracy(less than 0.1\% drop), therefore we narrow down the number of needed mask generators from 25 to 9, which correspond to the central $3\times3$ blocks that cover the main facial components. \emph{Stage} 3: After establishing our mask dictionary, we generate face samples with various random partial occlusions and calculate their corresponding FDMs using this dictionary. Then finetune the trunk CNN using these (face, mask) pairs with a small learning rate. This stage is designed for relieving the inconsistency between the real-value mask output by the mask generator and the final binarized version, so a few epochs are enough. 

\noindent \textbf{Testing.} In the testing stage, the similarity score is computed by the cosine distance of the fc features of two faces. The nearest neighbor classifier and thresholding are used for face identification and verification respectively. Considering the fact that, when recognizing an occluded face, we have lost the information from the occluded part of this face. Therefore it is necessary to also exclude this portion from the other faces comparing with it, to ensure that the similarity scores are computed based on equivalent information.

\noindent \textbf{Baseline Models.} Two baseline models are considered. The first one is the state-of-the-art face recognition model trained on CASIA-WebFace dataset. We will refer to it as \textit{Trunk CNN}. The second one has the same configuration with the first one but finetuned with synthetic occluded CASIA-WebFace dataset (average occluder area is 25\% of the face images), which will be referred to as \textit{Baseline}. 
\subsection{Ablation Study}
\label{sec:ablation}
\noindent \textbf{The Effect of $\tau$.} We conduct exploratory experiment to investigate the effect of $\tau$ used in binarization. By varying $\tau$ from 0 to 0.45, we evaluate our method on the AR dataset. The probe set contains faces with sunglasses and scarf occlusions, and the gallery set contains 1 clean face for every subject. The rank-1 identification accuracy is given in Table~\ref{tab:table1}. As $\tau$ being increased, the accuracy first rises up and then moves down as $\tau$ approaching 0.45. The best accuracy is achieved at $\tau=0.25$ and the performance is not highly sensitive to this threshold. 

\noindent \textbf{Mask Type.} To further explore the importance of binarization, we performed additional experiments with results shown in Table~\ref{tab:table2}. First, by comparing the ``Binary" and ``Soft weight", we see that ``Soft weight" noticeably decreases performance. We speculate that it's due to the excessive participation of features with very low mask values. Then we performed another experiment ``Soft+Binary" to remove those features with mask values lower than the threshold (setting these mask values to 0) while keeping mask values above threshold unchanged (rather than setting them as 1). This version achieved comparable performance to the Binary version. Obviously, the importance of binarization is to completely eliminate the noise by setting a feature element with a very low mask value to zero. At the same time, the binary mask is highly efficient in terms of both calculation and storage.
\begin{figure}[t]
\begin{center}
\includegraphics[width=0.8\linewidth]{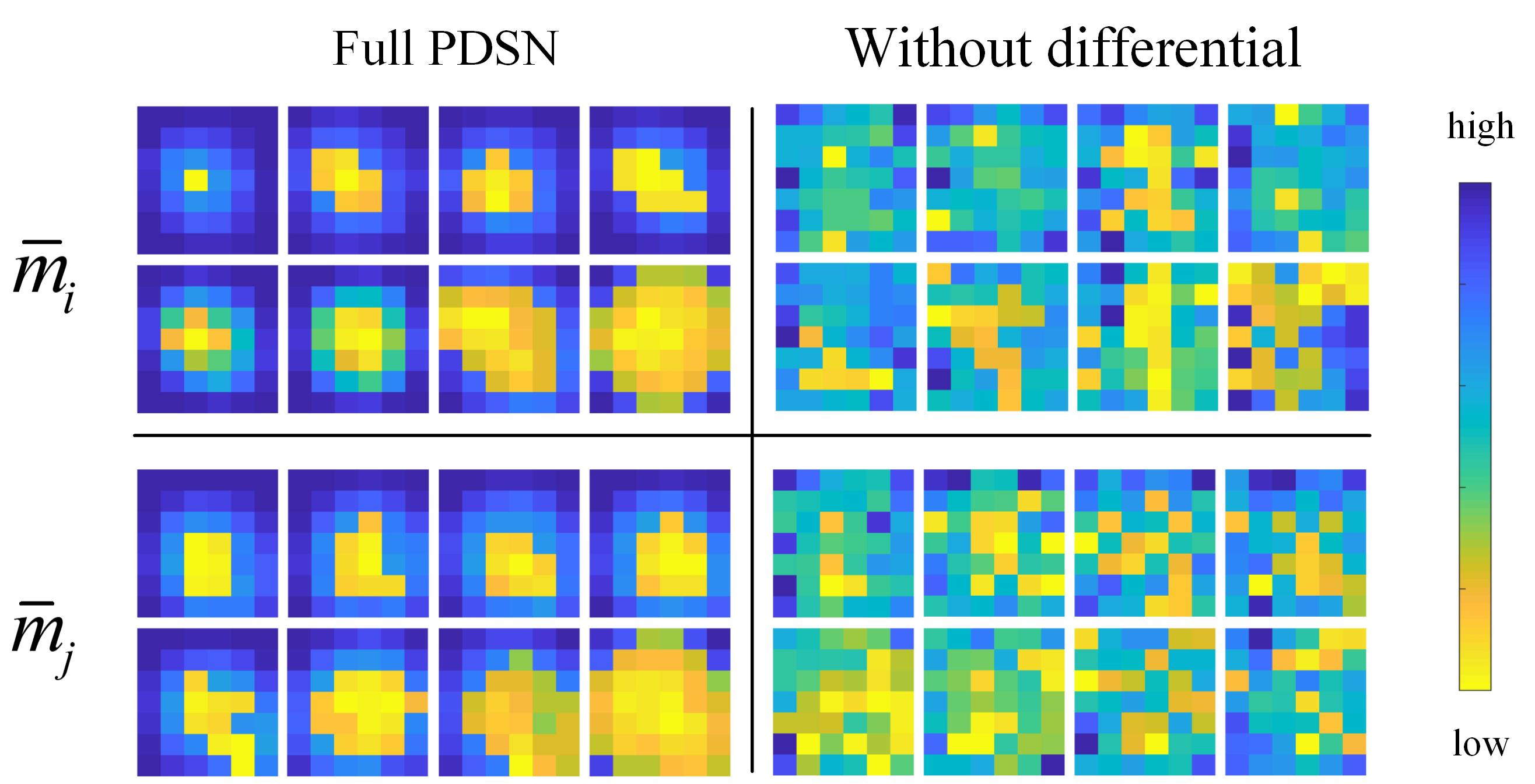} \\[0pt]
\end{center}
   \caption{Illustration of the mean masks learned by our full PDSN and only by classification loss. $\bar{m}_i$ corresponds to occlusion on the left eye block and $\bar{m}_j$ corresponds to occlusion on the nose block.}
\label{fig:figure7}
\end{figure}
\begin{table}[t]
\begin{center}
\begin{tabularx}{8.25cm}{|X<{\centering}|X<{\centering}|X<{\centering}|X<{\centering}|X<{\centering}|X<{\centering}|X<{\centering}|}
\hline
$\tau$ & 0 & 0.05 & 0.15 & 0.25 & 0.35 & 0.45 \\
\hline
Acc. & 95.84 & 97.29 & 97.36 & \textbf{98.26} & 97.98 & 97.92\\
\hline
\end{tabularx}
\end{center}
\caption[t]{Rank-1 identification accuracy(\%) comparison of different $\tau$ on AR dataset with sunglasses and scarf occlusions.}
\label{tab:table1}
\end{table}

\noindent \textbf{The Differential Supervision.} To investigate the importance of the differential input and pairwise loss. We set $\lambda$ in the loss function in Eq.~\eqref{equation5} to zero, and learn mask generators only from occluded face features. The mask dictionary is established with the same data and threshold $\tau$. The performance comparison is shown in Table~\ref{tab:table3}. The model trained with the pairwise supervision consistently outperforms the model that only trained with classification loss. In Figure~\ref{fig:figure7} we visualize several channels of the mean masks of the left eye and nose blocks respectively under these two conditions. With our full PDSN, the mask elements with much lower weights (highlighted part in Figure~\ref{fig:figure7}) could reflect the occlusion location in image space to some extent, which is reasonable since the top conv layer still preserves spatial information. While the mean masks generated by classification loss only are in chaos. As discussed above, the differential input and contrastive loss help the model concentrate on the feature elements that have been altered a lot by partial occlusions, while the classification loss alone is likely to also diminish feature elements that are affected by some other factors unrelated to occlusion. 
\subsection{Performance on LFW Benchmark}
LFW~\cite{lfw} is a standard face verification benchmark dataset under unconstrained conditions. We evaluate our models strictly following the standard protocol of unrestricted with labeled outside data and report the mean accuracy on the 6,000 testing image pairs.
\begin{table}[t]
\setlength{\abovecaptionskip}{0.cm}
\setlength{\belowcaptionskip}{-0.cm}
\begin{center}
\begin{tabularx}{8.25cm}{|p{1.8cm}<{\centering}|p{1.3cm}<{\centering}|p{1.7cm}<{\centering}|X<{\centering}|}
\hline
Mask type& Binary & Soft weight & Soft+Binary \\
\hline
\hline
Sunglasses & 98.19 & 96.67 & 98.19\\
Scarf & 98.33 & 97.22 & 99.03\\
\hline
\end{tabularx}
\end{center}
\caption[t]{The rank-1 identification accuracy (\%) in AR dataset Protocol 2. The results in the Protocol 1 have a similar conclusion.}
\label{tab:table2}
\end{table}
\begin{table}[t]
\begin{center}
\begin{tabularx}{8.25cm}{|p{1.8cm}<{\centering}|p{1.8cm}<{\centering}|p{1.5cm}<{\centering}|X<{\centering}|}
\hline
Differential & AR sunglass & AR scarf& MF1 occ \\
\hline
\hline
No & 95.97 & 97.92 & 54.80\\
Yes & 98.19 & 98.33 & 56.34\\
\hline
\end{tabularx}
\end{center}
\caption[t]{Rank-1 identification accuracy(\%) of our method with and without differential supervision information. ``MF1occ" refers to the occluded Facescrub probe set we synthesized.  }
\label{tab:table3}
\end{table}

As shown in Table~\ref{tab:table4}, the baseline model actually decreases the accuracy of the original trunk CNN by $0.52\%$ when it is trained to gain more robustness to partial occlusions because most of the face images in the LFW dataset are not occluded. This phenomenon is consistent with~\cite{increasing}, where performance encountered a drop when they tested a model that functions well for occluded objects on non-occluded objects. While our method can keep the performance of the trunk CNN since our design principle is just discarding those corrupted feature elements from comparison under partial occlusion condition, instead of forcing the trunk CNN to specifically accustom to partial occlusions.
\subsection{Performance on MegaFace Challenge1}
MegaFace Challenge~\cite{megaface} is a testing benchmark to evaluate the performance of face recognition algorithms at the million scale distractors. It contains a gallery set with more than 1 million face images. And the probe set consists of two datasets: Facescrub~\cite{facescrub} and FGNet. In this study, we use the Facescrub dataset as our probe set. The training set is viewed as small if it is less than 0.5M. We evaluate the basic trunk CNN, baseline model and our method under the small training set protocol on Challenge 1. The results are given in the ``MF1" column of Table~\ref{tab:table5}.

In order to test our method under partial occlusions, we synthesize the occluded Facescrub dataset. The occluding objects include sunglasses, mask, hand, eye mask, scarf, book, phone, cup, hat, fruit, microphone, hair, \etc., all of which are common objects in real-life that may appear on the face, and each type of occluding objects has several different images that are distinct from those used in training phase. The left four images in Figure~\ref{fig:figure6} show some examples. The results on this synthesized occluded Facescrub dataset are given in the ``MF1occ" column of Table~\ref{tab:table5}. Not surprisingly, a similar performance dropping on the original Facescrub probe set is observed for the baseline model. Compared to the baseline model, our method is superior on the occluded probe set without compromising the performance on the original probe set. 
\subsection{Performance on AR Dataset}
\label{sec:ARData}
We further evaluate our method through face identification experiments on the AR face database~\cite{AR} with real-life occlusions. The AR database contains 4,000 face images with different facial expressions, illumination conditions and occlusions from 126 subjects. There are mainly two kinds of testing protocols in the existing literature. \emph{Protocol} 1 refers to use more than 1 image per subject to form the gallery set (or training set). \emph{Protocol} 2 refers to use only 1 image per subject to form the gallery set. Images of sunglasses and scarf occlusions are used for testing. We evaluate our method under both protocols and the results are given in Table~\ref{tab:table6}. It is worth noting that the mask dictionary and the model are not finetuned with any AR face data at all, while other algorithms usually train with this dataset.
\begin{table}[t]
\begin{center}
\begin{tabularx}{8cm}{|p{2.2cm}<{\centering}|p{2cm}<{\centering}|p{1.2cm}<{\centering}|X<{\centering}|}
\hline
Method & Training Data & \#Models & Acc. \\
\hline
FaceNet~\cite{Tripletloss} & 200M & 1 & 99.63\\
DeepID2+~\cite{DeepID2+} & 2.6M & 3 & 98.95\\
CenterFace~\cite{centerloss} & 0.7M & 1 & 99.28\\
Baidu~\cite{baidu} & 1.3M & 1 & 99.13\\
SphereFace~\cite{sphereface} & 0.49M & 1 & 99.42\\
CosFace~\cite{cosface} & 5M & 1 & 99.73\\
ArcFace~\cite{arcface} & 0.49M & 1 & 99.53\\
\hline
\hline
Trunk CNN & 0.49M & 1 & 99.20\\
Baseline & 0.49M & 1 & 98.68\\
Ours. & 0.49M & 1 & \textbf{99.20}\\
\hline
\end{tabularx}
\end{center}
\caption[t]{Face verification(\%) on the LFW benchmark. ``\#Models" is the number of models used in the method for evaluation.}
\label{tab:table4}
\end{table}
\label{sec:Mega}
\begin{table}[t]
\begin{center}
\begin{tabularx}{8cm}{|p{2.7cm}<{\centering}|p{1cm}<{\centering}|p{1.5cm}<{\centering}|X<{\centering}|}
\hline
Methods & Protocol & MF1 & MF1occ \\
\hline
SIAT\_MMLAB & small & 65.23 & -\\
CenterFace~\cite{centerloss} & small & 65.49 & -\\
DeepSense & small & 70.98 & -\\
SphereFace~\cite{sphereface} & small & 72.73 & -\\
CosFace~\cite{cosface} & small & 77.11 & -\\
ArcFace~\cite{arcface} & small & 77.50 & -\\
FUDAN-CS\_SDS & small & 77.98 & -\\
\hline
\hline
Trunk CNN & small & 74.40 & 51.86\\
Baseline & small & 68.81 & 53.03\\
Ours. & small & \textbf{74.40} & \textbf{56.34}\\
\hline
\end{tabularx}
\end{center}
\caption[t]{Face identification accuracy(\%) on MegaFace Challenge 1. ``MF1occ" refers to the occluded Facescrub probe set. }
\label{tab:table5}
\end{table}
\label{sec:AR}
\begin{table}[t]
\begin{center}
\begin{tabularx}{8cm}{|p{2.4cm}<{\centering}|X<{\centering}|p{1.5cm}<{\centering}|p{1.4cm}<{\centering}|}
\hline
Methods & Protocol & Sunglass & Scarf \\
\hline
SRC\cite{SRC} & 1 & 87.00 & 59.50\\
NMR\cite{NMR} & 1 & 96.90 & 73.50\\
MLERPM\cite{mlerpm} & 1 & 98.00 & 97.00\\
SCF-PKR\cite{scfpkr} & 1 & 95.65 & 98.00\\
RPSM\cite{RPSM} & 1 & 96.00 & 97.66\\
MaskNet~\cite{MaskLearning} & 1 & 90.90 & 96.70\\
\hline
Trunk CNN & 1 & 98.19 & 99.72\\
Baseline & 1 & 99.58 & 99.86\\
Ours. & 1 & \textbf{99.72} & \textbf{100.0}\\
\hline
\hline
RPSM\cite{RPSM} & 2 & 84.84 & 90.16\\
Stringface\cite{stringface} & 2 & 82.00 & 92.00\\
LMA\cite{LMA} & 2 & 96.30 & 93.70\\
\hline
Trunk CNN & 2 & 95.14 & 96.53\\
Baseline & 2 & 96.67 & 96.39\\
Ours. & 2 & \textbf{98.19} & \textbf{98.33}\\
\hline
\end{tabularx}
\end{center}
\setlength{\belowcaptionskip}{0pt}
\caption[t]{Rank-1 face identification accuracy(\%) on the AR dataset with natural occlusions.}
\label{tab:table6}
\end{table}

Table~\ref{tab:table6} shows that our method can significantly improve the performance of the trunk CNN model on faces with real-life sunglasses and scarf occlusions. The superior performance of our method than the baseline model indicates that simply shrink the range affected by occlusion is definitely not enough, it is essential to eliminate the corrupted portion from the comparison because it brings information inconsistency. And our mask dictionary captures the intrinsic feature structure of the trunk CNN model, which generalizes well to other face samples.

\section{Conclusions}
In this paper, we proposed an occlusion robust face recognition approach with the pairwise differential siamese network (PDSN) that explicitly builds the correspondence between occluded facial blocks and corrupted feature elements. The competitive results on synthesized and realistic occluded face datasets demonstrate the superiority of the proposed approach over the state-of-the-arts, as well as the great generalization ability on general face recognition tasks.

{\small
\bibliographystyle{ieee_fullname}
\bibliography{egbib}
}

\end{document}